\documentclass[conference,a4paper]{IEEEtran}
\IEEEoverridecommandlockouts
\usepackage{amsmath,amssymb,amsfonts}
\usepackage{algorithmic}
\usepackage{url}
\usepackage{graphicx}
\usepackage{textcomp}
\usepackage{caption}
\usepackage{booktabs}
\usepackage{xcolor}
\usepackage{booktabs}
\def\BibTeX{{\rm B\kern-.05em{\sc i\kern-.025em b}\kern-.08em
    T\kern-.1667em\lower.7ex\hbox{E}\kern-.125emX}}
    
\begin{document}

\title{SpikeLogBERT: Energy-Efficient Log Parsing Using Spiking Transformer Networks\\
}

\author{
\begin{tabular}{cc}
\begin{tabular}{c}
Thuan Bui \\
\textit{Swinburne Vietnam, FPT University} \\
Hanoi, Vietnam \\
104486358@student.swin.edu.au
\end{tabular}
&
\begin{tabular}{c}
Duong Do \\
\textit{Swinburne Vietnam, FPT University} \\
Hanoi, Vietnam \\
105212305@student.swin.edu.au
\end{tabular}
\\[1em]
\begin{tabular}{c}
Tung Vu \\
\textit{Posts and Telecommunications Institute of Technology} \\
Hanoi, Vietnam \\
tung.vuson.hau@gmail.com
\end{tabular}
&
\begin{tabular}{c}
Duc-Tho Mai \\
\textit{Academy of Cryptography Techniques} \\
Hanoi, Vietnam \\
ductho-mai@actvn.edu.vn
\end{tabular}
\end{tabular}\\
Cong-Kha Pham\\
\textit{The University of Electro-Communications}\\ Tokyo, Japan\\
phamck@uec.ac.jp
}

\maketitle

\begin{abstract}
Log parsing is a fundamental step in automated log analysis, transforming raw system logs into structured event templates for downstream tasks such as anomaly detection and system monitoring. Existing log parsing methods range from rule-based and clustering-based approaches to neural models that learn semantic representations from log messages. However, neural approaches typically rely on dense matrix multiplications, which can result in high computational cost and energy consumption. This paper presents SpikeLogBERT, a spiking neural network framework for energy-efficient log parsing. The proposed model integrates a spiking transformer architecture with knowledge distillation from a BERT teacher model, enabling spike-driven computation while preserving semantic representation capability. By leveraging sparse spike activations and event-driven processing, the number of active operations during inference can be significantly reduced. As an initial benchmark study, experiments on the HDFS dataset demonstrate that SpikeLogBERT outperforms ANN-based neural log parsing models with a parsing accuracy of 0.99997, while reducing estimated theoretical energy consumption by up to $62.6\times$ under standard 45nm CMOS assumptions.
\end{abstract}


\section{Introduction}
Modern large-scale software systems continuously generate massive volumes of system logs recording critical events related to system execution, failures, and operational 
status - an essential data source for monitoring, anomaly detection, and root cause analysis. As cloud infrastructures and distributed systems continue to scale, automated log analysis has become a fundamental component in maintaining 
the reliability and operational stability of large-scale 
systems \cite{10301257}. Within this pipeline, log parsing 
plays a key role by transforming semi-structured log messages 
into structured event templates that can be effectively 
utilized by downstream log mining tasks.

A wide range of automated log parsing techniques has been proposed. Traditional approaches rely on rule-based heuristics, clustering strategies, or structural pattern 
matching to identify log templates. Recently, neural network approaches have been introduced to capture semantic relationships between log tokens using neural representations. 
Although these neural methods improve robustness and generalization, they typically rely on dense neural architectures involving extensive matrix multiplications, 
resulting in high computational cost and energy consumption. 
Such characteristics can limit their deployment in scenarios 
where log analysis must operate continuously or in 
resource-constrained environments.

Spiking Neural Networks (SNNs) have recently emerged as an energy-efficient alternative to conventional artificial neural networks. By using event-driven spike-based communication, SNNs enable sparse neuron activations and 
reduce the number of active operations during inference. Although spiking architectures have shown promising results in sequential modeling tasks, their application to system log analysis remains largely unexplored. Unlike general 
natural language processing tasks, log parsing operates on semi-structured messages with recurring template patterns and isolated variable parameters - a domain 
characteristic that naturally favors the sparse, event-driven computation of SNNs, yet has not been previously investigated in this context.

To address this gap, we introduce SpikeLogBERT, a spiking neural framework for energy-efficient log parsing. The proposed approach integrates BERT-based tokenization with 
a spiking transformer encoder to capture contextual relationships between log tokens under spike-driven computation. A knowledge distillation strategy is implemented, where a fine-tuned BERT model serves as a teacher network guiding the training of the spiking student model. The main contributions of this work are summarized 
as follows:

\begin{itemize}
    \item \textbf{SpikeLogBERT:} A spiking transformer with BERT-guided knowledge distillation for semantic log representation learning under spike-driven 
    computation, establishing a baseline for SNN-based log parsing research.
    
    \item \textbf{Theoretical energy benchmarking:} 
    A systematic evaluation comparing SNN and ANN neural log parsers under standardized 45nm CMOS energy assumptions, demonstrating up to $62.6\times$ estimated energy reduction while outperforming ANN baselines in parsing accuracy.
\end{itemize}

The rest of this paper is organized as follows. Section \ref{sec:related} reviews related work on log parsing techniques and recent developments in SNNs for energy-efficient computing. Section \ref{sec:proposed} presents the proposed SpikeLogBERT framework, including the spiking transformer architecture and the knowledge distillation strategy from a BERT teacher model. Section \ref{sec:experiments} describes the experimental setup and reports the performance evaluation results on benchmark log datasets. Finally, Section \ref{sec:conclusion} concludes the paper and discusses potential directions for future research.

\section{Related Works}\label{sec:related}

\subsection{Log Parsing Methods}

Log parsing aims to transform semi-structured log messages into structured event templates by separating constant patterns from variable parameters.  Early studies primarily relied on rule-based and heuristic-driven approaches, such as IPLoM~\cite{10.1145/1557019.1557154}, Spell~\cite{7837916}, and Drain~\cite{8029742}. These methods typically perform online clustering or structural matching based on token similarity and message patterns. While they are computationally efficient and require no labeled data, their performance is often sensitive to dataset-specific heuristics and may degrade when log formats evolve over time.

To improve robustness, learning-based log parsing approaches have been proposed to capture semantic relationships within log messages. Neural methods such as NuLog~\cite{10.1007/978-3-030-67667-4_8} and UniParser~\cite{10.1145/3485447.3511993} model logs using distributed representations, enabling improved generalization across heterogeneous log formats. More recently, prompt-based and large language model (LLM) approaches, including LogPPT~\cite{10172786} and DivLog~\cite{Xu2024}, have further improved parsing accuracy through pretrained language models. However, these neural approaches typically rely on large model architectures and dense matrix operations, which introduce substantial computational cost and may limit their deployment in resource-constrained environments.

\subsection{Spiking Neural Networks for Efficient Sequential Modeling}

SNNs have recently attracted increasing attention as an energy-efficient alternative to conventional artificial neural networks. Unlike traditional neural models that rely on continuous activations, SNNs employ event-driven spike-based communication, enabling sparse neuron activations and reducing active computations during inference.

While early SNN research primarily focused on computer vision tasks, recent studies have demonstrated their effectiveness in sequential and language modeling problems. Architectures such as the Spike-driven Transformer~\cite{10.5555/3666122.3668920} and SpikeBERT~\cite{LV2026108482} utilize ANN-to-SNN knowledge distillation to preserve semantic representation capability while reducing computational cost. Recently, spiking language models, including SpikeGPT~\cite{zhu2024spikegptgenerativepretrainedlanguage} and SpikingBert~\cite{10.1609/aaai.v38i10.28975}, have achieved competitive performance on natural language processing benchmarks while improving energy efficiency.

Despite these advances, existing SNN-based approaches have primarily focused on general language understanding tasks. The application of spiking neural architectures to system log analysis remains largely unexplored. In particular, automated log parsing requires efficient sequential processing of large volumes of log messages, making it a promising yet under-investigated domain for spike-driven computation.

\subsection{Energy Constraints in Neural Inference for Monitoring Systems}

Modern computing infrastructures increasingly rely on continuous monitoring and large-scale log analysis to maintain system reliability and operational stability. With the rapid growth of distributed systems and the Internet of Things (IoT), intelligent analytics is gradually shifting from centralized cloud platforms to edge computing to reduce latency, bandwidth usage, and privacy risks. Recent studies in Edge AI show that performing inference directly on resource-constrained devices can enable real-time decision-making while reducing communication and energy overhead~\cite{Witt2024,CONGNGUYEN2026111881}.

Nevertheless, deploying deep neural models in such environments remains challenging due to strict constraints on computation, memory, and power consumption. Consequently, lightweight and energy-aware learning frameworks have been actively explored to balance model accuracy with resource efficiency in embedded and monitoring systems~\cite{s25216629,liu2025starprivacypreservingenergyefficientedge}. These challenges highlight the growing need for energy-efficient intelligent analytics capable of processing large volumes of operational data under limited hardware resources.

\section{The Proposed Methods}\label{sec:proposed}
This section introduces the proposed SpikeLogBERT method. First, the core architecture of the event-driven Spiking Transformer is presented, detailing the substitution of dense matrix multiplications with sparse operations. Next, the multi-objective knowledge distillation pipeline is described, formulating how semantic comprehension is transferred from a continuous teacher network to the discrete spiking student.
\subsection{The Spiking Transformer Core Architecture}\label{sec:spiking}

\begin{figure}[htbp]
\centerline{\includegraphics[width=1\columnwidth]{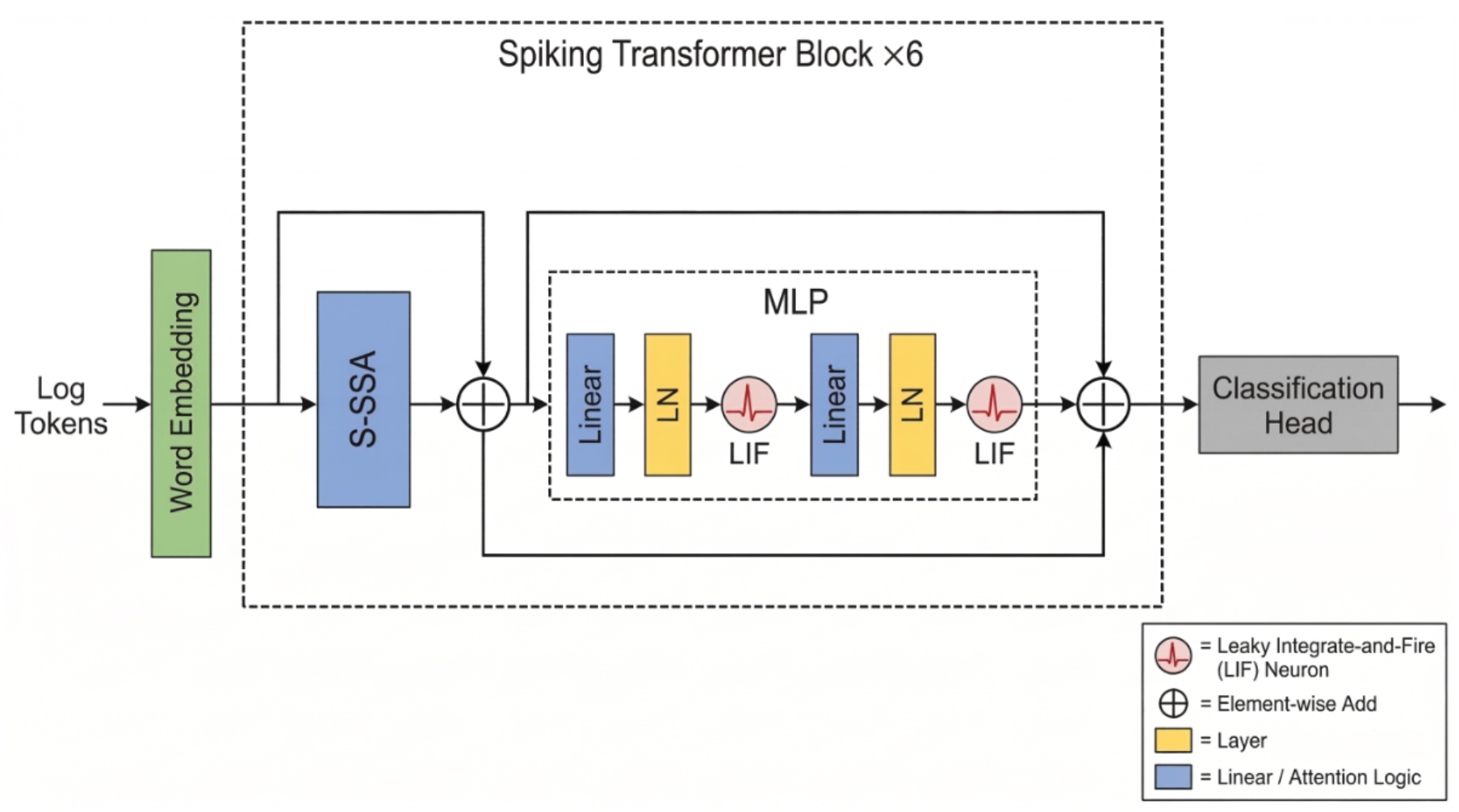}}
\caption{The architecture of Spiking Transformer Block.}
\label{fig:spikeformer}
\end{figure}

Unlike standard Artificial Neural Networks (ANNs) that rely on synchronous floating-point matrix multiplications, SpikeLogBERT employs an event-driven spiking neural network paradigm. As illustrated in Fig.~\ref{fig:spikeformer}, the foundation is the Spiking Transformer Block, processing binary spike sequences over $T$ discrete timesteps ($t \in \{1, \dots, T\}$), where $T$ denotes the total simulation length.

\begin{figure}[t]
\centerline{\includegraphics[width=0.5\textwidth]{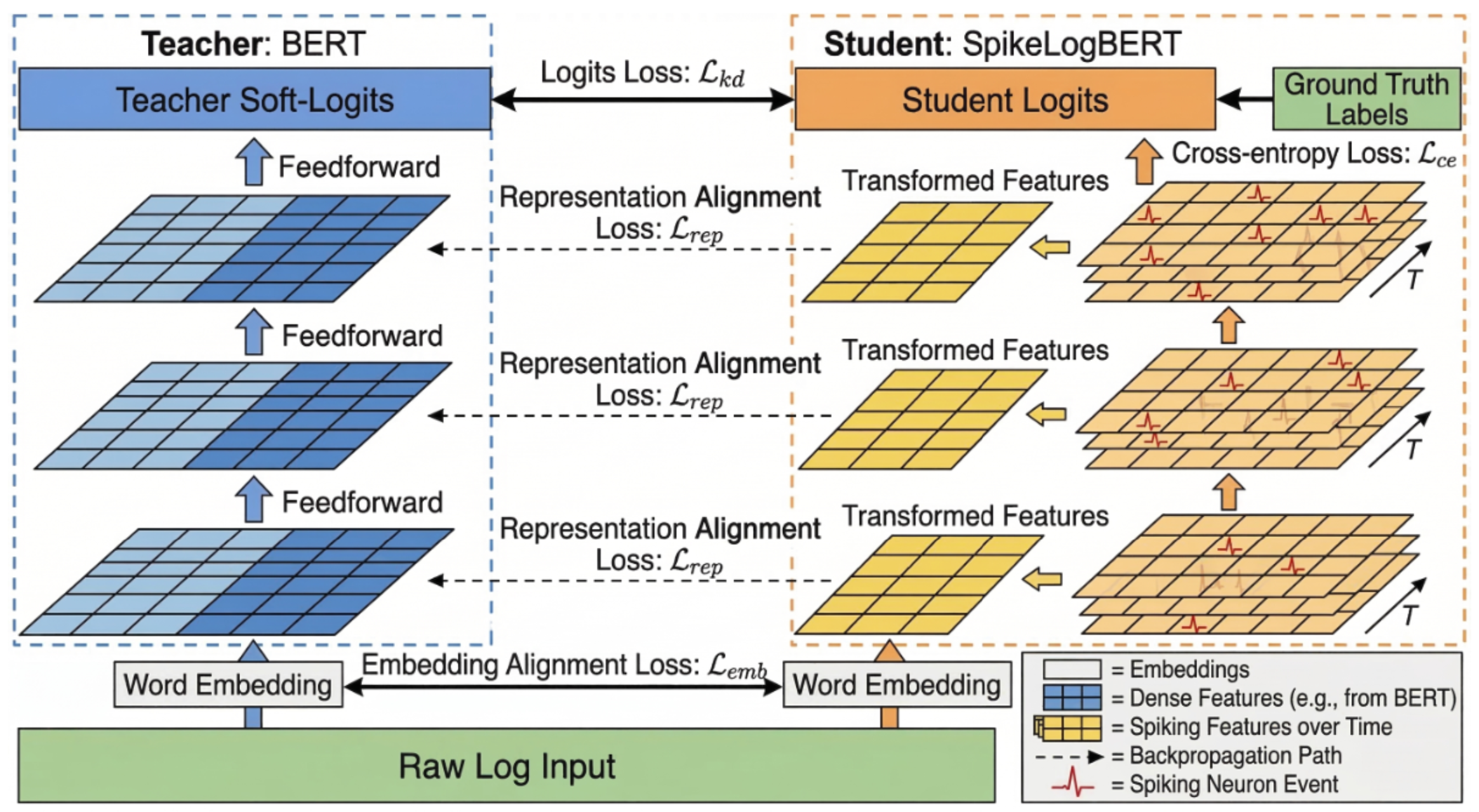}}
\caption{The Multi-Objective Knowledge Distillation Pipeline.}
\label{fig:distill}
\end{figure}

The fundamental non-linear activation unit is the Leaky Integrate-and-Fire (LIF) neuron, which sequentially accumulates input current $X^{(t)}$ into its membrane potential $U_{mem}^{(t)}$ at a given timestep $t$:
\begin{equation}
U_{mem}^{(t)} = \alpha U_{mem}^{(t-1)} + X^{(t)}
\end{equation}
where $\alpha \in [0, 1]$ is the leaky decay constant governing the persistence of past voltage. Once the membrane potential $U_{mem}^{(t)}$ exceeds a predefined firing threshold $U_{th}$, a discrete binary spike ($S^{(t)} = 1$) is emitted and the potential is reset; otherwise, the neuron remains dormant ($S^{(t)} = 0$).

The primary architectural innovation is the Spiking Self-Attention (SSA) module. While standard self-attention executes dense matrix multiplications requiring $\mathcal{O}(N^{2}D)$ operations (where $N$ is the sequence length and $D$ is the embedding dimension), the SSA mechanism restricts Query ($Q$), Key ($K$), and Value ($V$) tensors to binary spike formats. Consequently, the heavy matrix associative overhead is transformed into Sparse Accumulation (AC):
\begin{equation}
\mathrm{SSA}(Q, K, V) = \mathrm{LIF}\Big( \sum_{i=1}^{N} \sum_{j=1}^{N} Q_i \oplus K_j^T \Big) \otimes V
\end{equation}
where $Q_i$ and $K_j$ represent the binary spike vectors corresponding to the $i$-th and $j$-th tokens in the sequence ($i, j \in \{1, \dots, N\}$). Since both operands are binary, the inner product $Q_i \oplus K_j^T$ reduces to a logical AND followed by accumulation (AC), thereby avoiding floating-point multiplications entirely. Similarly, $\otimes$ denotes the subsequent sparse accumulation with the binary Value tensor $V$. By replacing MAC with AC operations, the hardware energy footprint is drastically reduced.

\subsection{Multi-Objective Knowledge Distillation}\label{sec:distill}

Despite the computational efficacy of SNNs, their optimization presents challenges. The discrete nature of LIF neurons obstructs continuous backpropagation, and training an SNN from scratch on semantic datasets often yields suboptimal convergence. To overcome these limitations, a multi-objective knowledge distillation pipeline guided by a pre-trained ANN (BERT-12L) is implemented, as conceptualized in Fig.~\ref{fig:distill}.

\subsubsection{Embedding Alignment}
To ensure an optimal latent vocabulary, the SNN embedding weights ($E_S$) are regularized to mirror the word embeddings of the BERT teacher ($E_T$) via a Mean Squared Error (MSE) loss:
\begin{equation}
\mathcal{L}_\mathrm{emb} = \mathrm{MSE}\Big(E_S, ~ E_T\Big)
\end{equation}
where $E_S, E_T \in \mathbb{R}^{\mathcal{V} \times D}$ are the respective embedding matrices for a vocabulary size $\mathcal{V}$ and embedding dimension $D$.

\subsubsection{Representation Alignment}
Layer-wise intermediate distillation enforces mapping the $l$-th student sequence representation to the activation of the $g(l)$-th teacher layer $H_T^{(g(l))}$. Specifically, with the student possessing $L_S=6$ layers and the teacher possessing $L_T=12$ layers, we adopt a uniform skip-layer mapping strategy defined as $g(l) = l \times (L_T / L_S) = 2l$. Because the SNN operates across the simulation timesteps $T$, the student's layer features are first temporally averaged. Let $\tilde{H}_S^{(t, l)}$ denote the hidden state output of the $l$-th Spiking Transformer Block at timestep $t$; the temporally averaged representation is then computed as $H_S^{(l)} = \frac{1}{T} \sum_{t=1}^{T} \tilde{H}_S^{(t, l)}$. A learnable linear transformation $W_{proj}^{(l)}$ then projects this averaged representation to match the teacher's hidden dimension, formulated via MSE across all $L_S$ student layers:
\begin{equation}
\mathcal{L}_\mathrm{rep} = \sum_{l=1}^{L_S} \mathrm{MSE}\Big(H_S^{(l)} W_{proj}^{(l)}, ~ H_T^{(g(l))}\Big)
\end{equation}

\subsubsection{Predictive Alignment}
The student model replicates the teacher's final decision boundary while learning from actual ground truth templates ($Y \in \{0, 1\}^C$, represented as one-hot encoded labels over $C$ log template classes). 

First, output soft-label distillation tethers the SNN logits ($Z_S \in \mathbb{R}^C$) to the teacher logits ($Z_T \in \mathbb{R}^C$). The discrepancy between predictive confidences is bounded using Kullback–Leibler (KL) Divergence:
\begin{equation}
\mathcal{L}_\mathrm{kd} = \tau_{kd}^2 \cdot \mathrm{KL}\Big(\sigma(Z_S/\tau_{kd}) ~||~ \sigma(Z_T/\tau_{kd})\Big)
\end{equation}
where $\sigma(\cdot)$ is the softmax function and $\tau_{kd}$ acts as a temperature scalar softening the target distribution. 

\begin{table*}[t]
\centering
\caption{Comparison of parsing accuracy on the HDFS dataset and theoretical energy consumption.}
\label{tab:results}

\begin{tabular}{lccccc}
\toprule
Model & Params (M) & FLOPs/SOPs (G) & Energy (mJ) & Energy Increase Compared to SpikeLogBERT & PA (\%) \\
\midrule
NuLog \cite{10.1007/978-3-030-67667-4_8} & \textbf{33.3} & 1.46 & 6.72 & $\times$8.2 & 99.8 \\
LogPPT \cite{10172786} & 108.3 & 11.17 & 51.38 & $\times$62.6 & 90.2 \\
\textbf{SpikeLogBERT (ours)} & 68.4 & \textbf{0.91} & \textbf{0.82} & -- & \textbf{99.997} \\
\bottomrule
\end{tabular}

\end{table*}

Second, standard cross-entropy loss ($\mathcal{L}_\mathrm{ce}$) computes the divergence between student logits $Z_S$ and true labels $Y$. The surrogate objective integrates all components into a unified backpropagation step:
\begin{equation}
\mathcal{L}_\mathrm{Total} = \lambda_1 \mathcal{L}_\mathrm{emb} + \lambda_2 \mathcal{L}_\mathrm{rep} + \lambda_3 \mathcal{L}_\mathrm{kd} + \lambda_4 \mathcal{L}_\mathrm{ce}
\end{equation}
where $\lambda_1, \lambda_2, \lambda_3, \lambda_4$ act as weight hyperparameters governing the magnitude of constituent losses. This integrated objective enables the Spiking Transformer to efficiently inherit state-of-the-art semantic comprehension capabilities.

\section{Experiments And Results}\label{sec:experiments}

This section evaluates the performance and computational characteristics of the SpikeLogBERT method. First, the dataset and experimental setup used for benchmarking are briefly described. Next, a theoretical energy consumption estimation is introduced to enable a fair comparison between SNN and ANN models. Finally, the parsing accuracy and energy efficiency of SpikeLogBERT are presented and discussed.

\subsection{Datasets and Experimental Setup}

Experiments were conducted on the HDFS log dataset \cite{10301257}. Parsing performance was evaluated using \textit{parsing accuracy} ($PA$), defined as the proportion of log messages whose templates are correctly identified. 

The proposed SpikeLogBERT student model was configured with $6$ Spiking Transformer layers ($L_S=6$), hidden dimension $768$ ($D=768$), and simulation length $16$ timesteps ($T=16$). The $12$-layer teacher BERT model ($L_T=12$) was fine-tuned for $16$ epochs with batch size $32$ and learning rate $\eta = 5\times10^{-5}$. The spiking student model was then trained for $30$ epochs under the same dataset configuration. All training was conducted on a workstation PC with an NVIDIA RTX 4090 GPU (24\,GB VRAM) and 64\,GB RAM.

\subsection{Theoretical Energy Consumption}

An important advantage of SNNs is their potential for energy-efficient inference on neuromorphic hardware. However, a direct comparison between SNN and ANN models is non-trivial, since the two paradigms rely on different computational primitives. Building upon the energy estimation principles established in prior spiking network studies \cite{10032591}, theoretical energy consumption was systematically evaluated by quantifying model computation into explicit operational costs.

For ANN-based models, computation was measured by floating-point operations ($\mathrm{FLOPs}$), corresponding to multiply-accumulate (MAC) operations, where $E_{\mathrm{MAC}}$ denotes the energy cost of one MAC operation. The theoretical energy consumption was evaluated as:
\begin{equation}
E_{\mathrm{ANN}} = E_{\mathrm{MAC}} \times \mathrm{FLOPs}
\end{equation}
For SNN-based models, computation was measured by synaptic operations ($\mathrm{SOPs}$), corresponding to spike-triggered accumulate (AC) operations, where $E_{\mathrm{AC}}$ denotes the energy cost of one AC operation. The theoretical energy consumption was evaluated as:
\begin{equation}
E_{\mathrm{SNN}} = E_{\mathrm{AC}} \times \mathrm{SOPs}
\end{equation}
By formulating these primitive operation counts, the total dynamic power can be accurately tracked. Under a $45\,\mathrm{nm}$ CMOS hardware assumption \cite{6757323}, the energy costs are standardized at $E_{\mathrm{MAC}}=4.6\,\mathrm{pJ}$ and $E_{\mathrm{AC}}=0.9\,\mathrm{pJ}$. This explicit conversion enables a rigorous and fair comparison of energy efficiency between SpikeLogBERT and traditional ANN parsing models.

\subsection{Results and Discussion}

Table~\ref{tab:results} presents the comparison between the SpikeLogBERT and representative neural log parsing models, including NuLog~\cite{10.1007/978-3-030-67667-4_8} and LogPPT~\cite{10172786}, on the HDFS dataset in terms of model size, 
computational cost, theoretical energy consumption, and parsing accuracy (PA). SpikeLogBERT achieves the highest 
parsing accuracy with $PA = 99.997\%$, outperforming NuLog 
($PA = 99.8\%$) and LogPPT ($PA = 90.2\%$). Although NuLog has the 
smallest parameter size, the proposed model maintains higher accuracy 
while preserving a relatively compact architecture.
More importantly, SpikeLogBERT demonstrates a substantial advantage in 
increase of approximately $8.2\times$ for NuLog and $62.6\times$ for 
LogPPT compared with SpikeLogBERT.

The significant reduction in energy consumption is primarily attributed 
to the sparse event-driven computation of SNNs, where synaptic operations are triggered only when spike events occur. As a result, 
SpikeLogBERT maintains SOTA parsing accuracy while 
substantially reducing theoretical energy consumption compared with ANN models.

\section{Conclusion}\label{sec:conclusion}
This paper presented SpikeLogBERT, a spiking neural network framework for energy-efficient log parsing that integrates a spiking transformer architecture with a multi-objective knowledge distillation pipeline, enabling robust semantic 
representation learning from a pre-trained BERT teacher. 
Experimental evaluations on the HDFS dataset demonstrate that SpikeLogBERT outperforms representative ANN-based neural log parsers, achieving a parsing accuracy of 
$0.99997$ while reducing the estimated theoretical energy consumption by up to $62.6\times$ under standard 45nm CMOS assumptions. These results establish SpikeLogBERT as a promising baseline for SNN-based log parsing research, 
highlighting the potential of spike-driven computation for energy-aware log analysis in resource-constrained environments. Future work will extend this evaluation to 
diverse cross-system datasets and validate the estimated energy gains through direct deployment on physical neuromorphic hardware.
\section*{Acknowledgment}
This work was supported by JST NEXUS, Japan Grant 
Number JPMJNX25D4.

\clearpage
\bibliographystyle{IEEEtran}
\bibliography{ref}

\end{document}